%% file: main.tex
\documentclass{article} 
\usepackage{style,times}

\input{math_commands.tex}

\usepackage[pagebackref=true,colorlinks,
            linkcolor=blue!60!black,
            citecolor=blue!70!black,
            urlcolor=blue!50!black]{hyperref}
\usepackage{url}
\usepackage{comment}

\usepackage{xspace}
\usepackage{xcolor}
\usepackage[table]{xcolor}
\usepackage{graphicx}
\usepackage{subcaption}
\usepackage{amsmath}
\usepackage{booktabs}
\usepackage{caption,multirow}
\usepackage{siunitx}
\usepackage{wrapfig}
\usepackage{tabularx}
\usepackage[normalem]{ulem}

\usepackage{pifont}

\newcommand{\name}{DUCHESS\xspace}

\newcommand{\rana}[1]{\textcolor{orange}{[Rana: #1]}}

\title{Intra-Request Branch Orchestration for\newline Efficient LLM Reasoning}

\author{Weifan Jiang, Rana Shahout, Yilun Du, Michael Mitzenmacher \& Minlan Yu\\
Harvard University\\
Boston, MA 02163, USA
}

\begin{document}

\maketitle

\begin{abstract}

 LLMs increasingly rely on inference-time reasoning algorithms such as chain-of-thought and multi-branch reasoning to improve accuracy on complex tasks. These methods, however, significantly increase token usage (cost) and per-request latency.
Prior work has primarily focused on reducing token usage, often at the expense of accuracy, while overlooking other latency factors.

We present \name, an LLM serving system that reduces computational cost and latency without sacrificing accuracy through intra-request branch orchestration guided by predictions.
Within each request, \name predicts branch correctness with
a lightweight linear probing model over LLM layer activations. The orchestration policy uses these predictions to decide whether to terminate a branch early, duplicate an existing branch, or continue exploring a branch.
When handling multiple requests, \name can further reduce latency by prioritizing easier reasoning tasks, when request complexity can be estimated from the prompt.

Experiments on three reasoning benchmarks show that \name consistently improves the token–accuracy Pareto frontier, reducing token usage by 42–63\% at matched accuracy compared to self-consistency. For request serving with vLLM, \name reduces mean, median, and tail latencies by 57-81\%, 58-85\%, and 52-84\% with First-Come-First-Served (FCFS) scheduling across three datasets, compared to self-consistency. At higher request rates, scheduling jobs by increasing predicted difficulty reduces latency further over FCFS.

\end{abstract}

\input{sections/introduction}

\input{sections/background}

\input{sections/method}

\input{sections/evaluation}

\input{sections/related_works}
\input{sections/conclusion}



\bibliography{references.bib}
\bibliographystyle{style.bst}

\appendix
\input{sections/appendix}

\end{document}

%% file: math_commands.tex
\usepackage{amsmath,amsfonts,bm}

\def\secref#1{section~\ref{#1}}

\def\eqref#1{equation~\ref{#1}}

\def\1{\bm{1}}

\DeclareMathAlphabet{\mathsfit}{\encodingdefault}{\sfdefault}{m}{sl}
\SetMathAlphabet{\mathsfit}{bold}{\encodingdefault}{\sfdefault}{bx}{n}



%% file: sections/introduction.tex
\section{Introduction}

Large Language Models (LLMs) are widely applied across domains, including math and science problem-solving~\citep{lewkowycz2022solving}, coding and program analysis~\citep{jiang2025surveyonllmforcode}, logical deduction~\citep{creswell2023selectioninference}, and decision-making~\citep{yao2023tree}. Solving these tasks often requires \emph{reasoning algorithms}, which extend beyond one-shot prediction. Reasoning algorithms~\citep{wang2023selfconsistency, wei2022chainofthought, yao2023tree, wu2024scaling, snell2025scaling} vary in implementation but share the goal of improving LLM reasoning at inference time. They enable models to reflect on their outputs, explore alternative reasoning paths, and produce more accurate answers. Among these methods, multi-branch reasoning~\citep{wang2023selfconsistency} has become a widely used approach. The model generates multiple candidate reasoning paths (branches) for a request and aggregates their outcomes to produce the final answer. While individual branches may use chain-of-thought~\citep{wei2022chainofthought} reasoning, the defining feature of multi-branch reasoning is the exploration of multiple paths in parallel.

While reasoning algorithms improve accuracy, they also introduce substantial computational overhead. Long chains of thought and multiple branches increase token generation per request, and each additional token extends processing time. As LLMs are increasingly deployed in interactive and large-scale applications, these inefficiencies directly impact scalability and user experience. Building efficient LLM reasoning systems, therefore, requires balancing cost, latency, and accuracy, which gives rise to two main challenges.

The first challenge is the cost–accuracy tradeoff: pruning output tokens reduces inference cost but can harm accuracy. For example, terminating a branch’s reasoning too early~\citep{zhang2025reasoningmodelsknowtheyre, afzal2025knowingbeforesaying, yang2025dynamicearlyexitreasoning, liu2025answerconvergencesignalearly, nvidia2025tensorrt} risks missing the correct answer, while stopping too late wastes tokens. In multi-branch reasoning, accuracy also depends on how different branches interact, yet existing methods often treat branches independently and optimize only for cost~\citep{hassid2025dontoverthinkitpreferring, li2024escape, fu2024efficient}. Beyond this cost–accuracy tradeoff, a second challenge arises in maintaining low latency under concurrent serving, where multiple reasoning requests compete for shared resources. Latency depends not only on the number of generated tokens but also on system-level factors such as queuing delays, parallel execution efficiency, and straggler branches that determine request completion time. Methods that reduce token counts, such as adaptively generating branches~\citep{li2024escape}, can be less parallelizable and therefore slow down throughput. Other approaches rely on complex orchestration—using an auxiliary model~\citep{jin2023s3, leviathan2023speculative} or reinforcement learning~\citep{aggarwal2025l1controllinglongreasoning, dai2025sgrpoearlyexitreinforcement, hou2025thinkprune, lou2025adacotparetooptimaladaptivechainofthought} which improves token efficiency but introduces additional overhead that may increase end-to-end latency.

In this paper, we present \name, an LLM serving system that reduces cost and latency while maintaining accuracy.
Given a reasoning request, \name orchestrates its branches to complete it as efficiently as possible. It applies an intra-request branch orchestration policy guided by predictions derived from LLM layer activations. These predictions estimate the correctness of each branch and inform three possible branch-level actions:
\begin{itemize}
    \item 
Early termination: stop a branch once it is likely to already yield a correct answer, avoiding unnecessary token generation.
    \item 
Selective branch-out: duplicate a promising branch to increase coverage while reusing its past progress, rather than starting from scratch.
    \item 
Continuation: keep generating tokens for branches that are still uncertain, giving them additional steps to potentially reach a correct answer.
\end{itemize}
These branch-level decisions reduce token usage with minimal accuracy loss. Additionally, the system terminates the entire request once either a consensus among enough branches is reached or a sufficient number of answers are collected, preventing straggling branches from delaying completion.



At the request-pool level, \name can optionally apply a lightweight scheduling step before intra-request orchestration. If request difficulty can be estimated from the prompt, \name prioritizes easier requests first, following the intuition of Shortest Job First (SJF) and prior works on LLM serving systems~\citep{shahout2025dont, shahout2024fast, fu2024efficient, mitzenmacher2025queueing}. Otherwise, it defaults to First-Come-First-Served (FCFS). This scheduling is not central to our approach but can offer additional latency benefits when complexity information is available.

We evaluate \name using three popular reasoning benchmarks datasets: GSM8K~\citep{cobbe2021gsm8k}, MMLU~\citep{hendrycks2021test}, and MATH~\citep{hendrycks2021math}, with DeepSeek-R1-Distill-Llama-8B~\citep{guo2025deepseek} as the model. Our results demonstrate that \name efficiently reduces reasoning cost and latency without trading accuracy. \name dominates the cost-accuracy Pareto frontier in all of our experiments, reducing token usages by 42\%\text{-}63\% at matched accuracy compared to Self-consistency~\citep{wang2023selfconsistency}, a popular multi-branch reasoning algorithm. For request serving with vLLM~\citep{kwon2023pagedattention}, \name reduces mean, P50, and P95 latencies by 57-81\%, 58-85\%, and 52-84\% on three datasets using First Come First Served (FCFS). Furthermore, on the MATH dataset with difficulty-labeled requests, at higher arrival rates applying complexity-aware scheduling reduces \name mean latency by up to 29.7\% and 34.7\% compared to FCFS, using predicted and actual difficulty labels, respectively.


%% file: sections/background.tex
\section{Background and Related Works}

\textbf{Single CoT token reduction.} Recent works have studied pruning a single CoT sequence to reduce token usage, and/or avoid accuracy drops caused by overthinking. \citep{zhang2025reasoningmodelsknowtheyre, afzal2025knowingbeforesaying} predicts future CoT correctness based on current LLM hidden states;~\citep{yang2025dynamicearlyexitreasoning} identifies transit points in reasoning processes such as ``wait'' for potential early exit opportunities.~\citep{liu2025answerconvergencesignalearly, fu2024efficient} relies on intermediate answer consistency for CoT convergence. Overall, these works do not consider cross-branch dependencies within a request, which can compromise the optimality of cost-accuracy tradeoffs.

\textbf{Multi-branch token reduction.} In contrast to the above, some works jointly optimize all branches within the request, which usually delivers more optimal cost-accuracy tradeoffs due to reasoning awareness but sometimes overlook latency requirements. Incremental self-consistency methods (e.g., ESC~\citep{li2024escape}, RASC~\citep{wan2025reasoningaware}, Adaptive Consistency~\citep{aggarwal2023lets}) generate branches in small batches until certain confidence conditions. These works can achieve good token-accuracy tradeoffs, but are unable to be sufficiently parallelized. Some use complex solutions (e.g., another transformer/LLM~\citep{jin2023s3, leviathan2023speculative} or reinforcement learning methods~\citep{aggarwal2025l1controllinglongreasoning, dai2025sgrpoearlyexitreinforcement, hou2025thinkprune, lou2025adacotparetooptimaladaptivechainofthought}) to orchestrate request serving or reasoning progress at high additional overhead.~\citep{hong2025slimsc} prunes reasoning branches based on similarities.

%% file: sections/method.tex
\section{Method}

\begin{figure}[t]
  \centering
  \includegraphics[width=\linewidth]{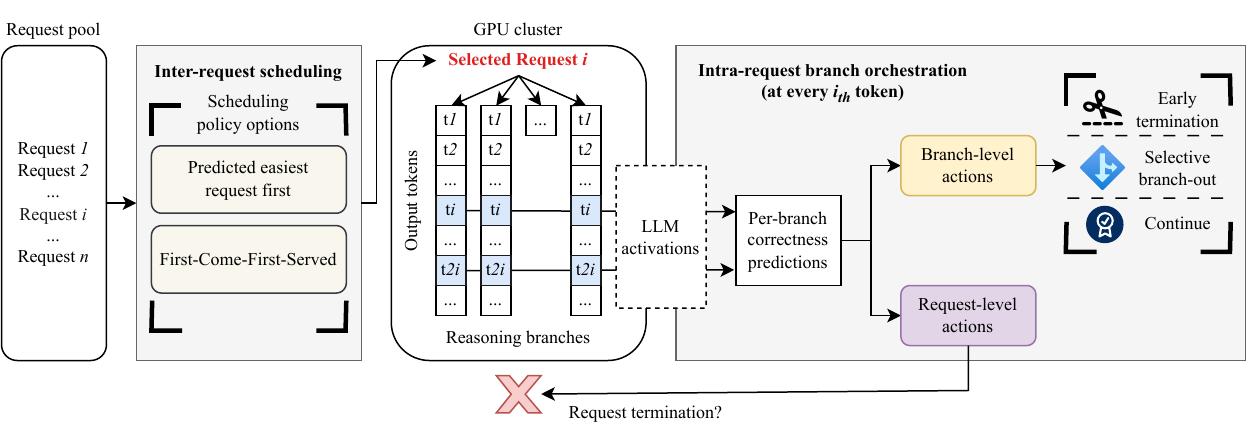}
  \caption{\label{fig:sys-overview} \textbf{\name overflow.}
  \name schedules requests by prioritizing easier ones when complexities can be estimated from the prompt, otherwise defaulting to First-Come-First-Served.
  To process a single request, at every $i_{th}$ token, \name predicts branch correctness from LLM activations and decides: at the branch level, whether to terminate, branch out, or continue. 
  At a request level, the decision is made on whether to terminate the entire request once either a suitable consensus is reached or a sufficient number of answers are collected.}
\end{figure}

\name reduces the computational cost and latency of long CoT multi-branch reasoning requests without sacrificing accuracy. At each step, \name selects a request from the pool and completes it as quickly as possible to reduce wait time. Within each request, \name applies an intra-request branch orchestration policy: after every $i^{th}$ token, it uses a lightweight linear probing model on transformer activations to estimate per-branch correctness probabilities. Based on these predictions, \name determines both branch-level and request-level actions. At the branch level, it assigns each branch one of three actions: 1) early termination, 2) selective branch-out, or 3) continuation. This policy avoids wasted computation on low-value branches, producing accurate answers with fewer tokens while preserving branch-level parallelism to reduce latency.
At the request level, since multi-branch reasoning relies on majority voting, \name terminates an entire request once either a majority consensus is reached or a sufficient number of answers are collected, preventing stragglers from delaying completion.

When scheduling across multiple requests, \name can optionally apply a complexity-aware policy that prioritizes easier requests if their difficulty is known or can be estimated from the prompt. Following the intuition of Shortest Job First (SJF), this step helps reduce average latency across requests. Figure~\ref{fig:sys-overview} illustrates \name's workflow.




\subsection{Per-branch correctness prediction.}

In this section, we describe our method for estimating branch correctness from transformer activations. Specifically, we employ a linear probing model to predict per-branch correctness, i.e., the probability that a correct answer can be produced from the branch’s current CoT steps.



\noindent\textbf{Problem definition.} 
Given a prompt $p$ and a partially decoded branch consisting of tokens $t_1, t_2, \dots, t_k$, 
we aim to estimate whether the LLM can already produce a correct answer based on this partial sequence. 
If so, we terminate the branch and apply CoT probing~\citep{fu2024efficient}, which queries the LLM 
for a direct answer using the existing tokens with the following probing prompt:
\[
p + t_1 t_2 \dots t_k + \texttt{``**Final Answer**\textbackslash n\textbackslash n\textbackslash[ \textbackslash boxed\{''}
\]
The final segment instructs the LLM to output a direct answer conditioned on the tokens $t_1 \dots t_k$.

In this work, we use CoT probing to extract answers from incomplete reasoning sequences. This technique requires only a small token budget (e.g., about 10 tokens) since the probed output is a short direct answer. \citep{fu2024efficient} proposes applying CoT probing every 32–64 tokens and using certainty, whether probed answers remain stable across consecutive intervals, as a proxy for correctness.
\name{}, in contrast, predicts correctness probabilities directly from the transformer activations of the last decoded token $t_k$. Since $t_k$ encodes the reasoning progress up to that point, its activations provide informative features for correctness prediction. These activations are produced during normal decoding, so our method adds no extra collection or processing overhead.

\begin{figure}[t]
  \centering
\includegraphics[width=\linewidth]{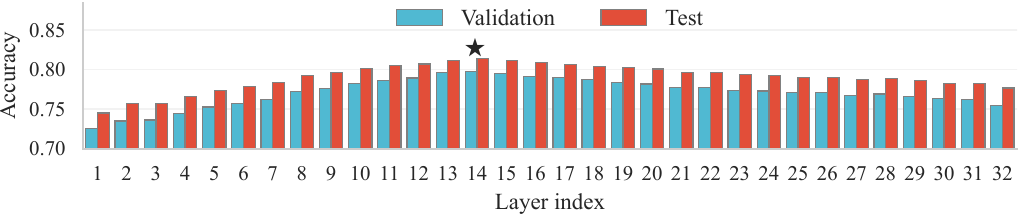} 
\vspace{-20pt}
\caption{\label{fig:pred_acc_by_layer} \textbf{LLM activations produce accurate correctness predictions.} The validation and test accuracies of the proposed MLP trained on each LLM transformer layer, using DeepSeek-R1-Distill-Llama-8B model and GSM8K dataset. Layer 14 yields highest validation and test accuracies.}

\end{figure}


\noindent\textbf{Predictor design.} We design the predictor as a multi-layer perceptron (MLP) with two fully connected hidden layers. The input is the activation of token $t_k$ from one of the LLM’s embedding layers. Although using more layers could improve accuracy, it would also increase computational overhead; therefore, we use a single layer for simplicity. The MLP outputs a single logit representing the probability that a CoT probing after token $t_k$ would yield a correct answer. We implement and train the predictor in PyTorch~\citep{paszke2019pytorch}, with minor dataset-specific hyperparameter adjustments based on validation performance. Detailed model architectures are provided in \ref{sec:eval-system-setting}.

\noindent\textbf{Data collection and predictor training.} We construct the training set from \textless prompt, groundtruth \textgreater\xspace pairs. For each prompt, we generate 10 LLM responses, each capped at 4,096 tokens. From every response, we apply CoT probing at intervals of $i=16$ tokens to extract intermediate answers. Smaller intervals produce more training samples but increase overhead; empirically, $i=16$ provides accurate predictions at a manageable cost. We assign a positive label if the intermediate answers are correct for three consecutive probes starting at $t_i$, which filters out ``lucky guesses'' where correctness fluctuates. The MLP uses the activation of $t_i$ as input features and the correctness label as the target. We train the MLP with cross-entropy loss and the AdamW optimizer.
We evaluate this approach on datasets such as GSM8K, MMLU and MATH, and train the predictor separately on each dataset’s training split.

Figure~\ref{fig:pred_acc_by_layer} reports validation and test accuracy when training on activations from different LLM layers, using the GSM8K dataset and DeepSeek-R1-Distill-Llama-8B model. The middle layers yield the highest accuracy, peaking at layer 14, consistent with prior findings that middle layers capture the richest information in a transformer~\citep{shahout2025dont, skean2024does}.

\subsection{Intra-request branch orchestration}

\name orchestrates a single request using predictions of branch correctness. 
After every $i=16$ tokens per branch, it applies the activation-based MLP to estimate the probability that each branch can yield a correct answer.
Based on these predictions, \name assigns branch-level actions and, when applicable, request-level actions.

Smaller intervals $i$ provide finer-grained monitoring but increase prediction overhead.
In practice, we found $16 \leq i \leq 80$ to be a good tradeoff.

\paragraph{Branch-level actions}

At the branch level, \name supports three actions based on predicted correctness: (1) early termination, (2) selective branch-out,
and (3) continuation.

\noindent\textbf{1. Early termination.} 
\name terminates a branch if its predicted correctness exceeds a threshold $\tau$ for $S$ consecutive rounds. 
We set $\tau$ to the 70-80th percentile of validation predictions and $S=2$ to prevent premature termination. 
When a branch is terminated, \name applies CoT probing to extract a final answer from its current progress. 
If the branch ends naturally (EOS token), its final answer is retained without probing.

\noindent\textbf{2. Selective branch-out.} 
Terminated branches free GPU slots. 
Rather than start new branches from scratch, which risks stragglers, \name forks a new branch from an existing one, reusing its CoT progress and KV cache to reduce overhead. 
When there are $r$ active branches with correctness probabilities $p_1, \dots, p_r$, \name samples according to a rescaled distribution:
\vspace{-0.05in}
\[
\hat{p}_{j} = \frac{p_j^{1/\lambda}}{\sum_{k=1}^r p_k^{1/\lambda}}, \quad 1 \leq j \leq r,
\]
where $\lambda$ is a temperature parameter. 
Smaller $\lambda$ favors high-confidence branches, while larger $\lambda$ increases the chance of duplicating lower-confidence ones. 
We tune $\lambda$ on validation data.

\noindent\textbf{3. Continuation.} 
Branches that are not terminated continue decoding for another $i$ tokens before being re-evaluated.

\vspace{-0.12 in}

\paragraph{Request-level action}

Since multi-branch reasoning relies on majority voting, \name terminates the entire request once enough answers are available. 
We propose two conditions: (1) at least $\alpha \cdot c$ identical answers are collected, or (2) at least $\beta \cdot c$ answers are collected, where $0 < \alpha < \beta < 1$ and $c$ is the number of parallel branches. 
The first condition captures fast consensus, while the second ensures sufficient coverage. 
Smaller $\alpha$ and $\beta$ values lead to more aggressive termination at the cost of potential accuracy loss. 





\noindent\textbf{Parallelism.} 
Additionally, \name maintains up to $c$ active branches at all times. In practice, $c$ can be set based on maximum batch size allowed by hardware. This design ensures that GPU resources are fully utilized and prevents idle slots, thereby reducing per-request latency.

\vspace{-0.05in}

\subsection{\label{sec:inter-request} Inter-request complexity-aware scheduling.}  

\vspace{-0.05in}

In pooled serving, prioritizing requests with shorter runtimes reduces mean latency. For LLM requests, runtimes must be predicted because autoregressive decoding is non-deterministic. Existing work~\citep{shahout2025dont, jin2023s3, zheng2023response} focuses on predicting output token lengths; however, actual runtimes also depend on algorithmic factors (e.g., early termination) and system factors (e.g., prefix caching), making them harder to estimate.  

We observe that \name naturally completes easier requests faster. For example, in the MATH dataset~\citep{hendrycks2021math}, where requests are labeled with difficulty levels from 1 (easy) to 5 (hard), as difficulty increases, so does the average time \name spends per request: 13.8, 18.4, 25.8, 33.6, 47.4 for levels 1-5 (using 10 branches).
Motivated by this, we introduce an optional complexity-aware scheduling policy: after completing one request, \name selects the easiest available request from the queue, provided difficulty labels are known. Otherwise, \name defaults to FCFS.

While the MATH dataset has labels, generally difficulty labels will need to be predicted. We therefore replaced the exact labels in the MATH dataset with predicted labels (using the original labels for training).
Leveraging the informativeness of LLM activations, we developed a simple MLP model with three hidden layers for request complexity prediction using the MATH dataset. The input to the MLP is the activation of the first decoded token (i.e., after prefilling stage) from layer 14 of R1-Distill-Llama, and the output is the probability distribution across 5 complexity levels.

When using predicted complexities to schedule requests, \name first performs prefilling for all newly arrived requests, since the predictor relies on prefilling activations as input. Although the prefilling increases request queueing delays, its cost is insignificant relative to decoding, ensuring that \name still delivers end-to-end latency reductions.

We show in Appendix~\ref{sec:complexity-prediction-acc} the confusion matrix for predicted request complexities. In summary, we observe our predictor is able to differentiate easier and harder requests well, which has been shown to reduce queueing delays \citep{mitzenmacher2021queues, mitzenmacher2025queueing}.

%

%% file: sections/evaluation.tex
\section{Evaluation}

\vspace{-0.1 in}
\subsection{Experimental settings.}


\noindent\textbf{Datasets and model.} We evaluate \name on three popular reasoning benchmarks: GSM8K~\citep{cobbe2021gsm8k}, MATH~\citep{hendrycks2021math}, and MMLU~\citep{hendrycks2021test}. We use the official train, validation, and test splits of the MATH dataset. For the other datasets, we use their official test splits, and randomly partition the train split into train and validation sets due to the lack of official validation splits, following approximate 70-10-20 ratios. We use DeepSeek-R1-Distill-Llama-8B model~\citep{guo2025deepseek} as it natively generates CoT-style responses without the need for prompting.



\noindent\textbf{Metrics.} Our primary metrics are: \textbf{(1) Cost-accuracy tradeoff.} We vary the maximum number of branches per request $5 \leq c \leq 10$, close to settings used by related works~\citep{hassid2025dontoverthinkitpreferring}. For each $c$, we record inference cost (i.e., the average number of output tokens generated per request) and the average accuracy across all requests. The max token size per branch is capped at 4,096. Our goal is to achieve better Pareto optimality, that is, using fewer tokens at matched accuracy, and achieving higher accuracy at matched tokens, compared to existing systems. \textbf{(2) Mean/P50/P95 latencies and Time-To-First-Tokens (TTFT).} We randomly generate request arrival schedules following Poisson distributions with an average rate of $2$ queries per minute (QPM) for GSM8K and MMLU, and $1$ for MATH. These rates are determined based on the upper quantiles of default self-consistency processing time per request at $c=10$ (26.5, 30.5, and 64.0 seconds, respectively), thereby reflecting realistic loads in a single-server setting. We report the mean, P50, and P95 latencies (time between request arrival and completion) and TTFTs (time between request arrival and first decoded token): the former reflects end-to-end runtime, whereas the latter emphasizes queueing delays. We use vLLM v0.7.3~\citep{kwon2023pagedattention} with chunked prefilling and prefix caching enabled, on a testbed with one NVIDIA-A100 80GB GPU and 64 AMD EPYC 7313 16-Core Processors.


\noindent\textbf{Baselines.} We compare \name to: \textbf{(1) Default self-consistency}~\citep{wang2023selfconsistency} with no token-saving efforts. \textbf{(2) Dynasor}~\citep{fu2024efficient} for single branch early termination, which each branch is prompted to generate an intermediate answer at fixed token intervals, and early terminated if a sufficient number of consecutive consistent intermediate answers are observed. \textbf{(3) Short-m@k}~\citep{hassid2025dontoverthinkitpreferring} for request-level early termination, which generates $k$ branches in parallel and terminates the whole request after the first $m$ out of $k$ finish, handling straggler effects at the cost of fewer final answers. For (2) and (3), we use the default settings from the original papers if possible, and only adjust them to achieve comparable accuracy levels in certain cases.

Our full experimental settings are recorded in \ref{sec:eval-system-setting}.

\subsection{Evaluation results.}

\begin{figure}[t]
  \centering
  \includegraphics[width=\linewidth]{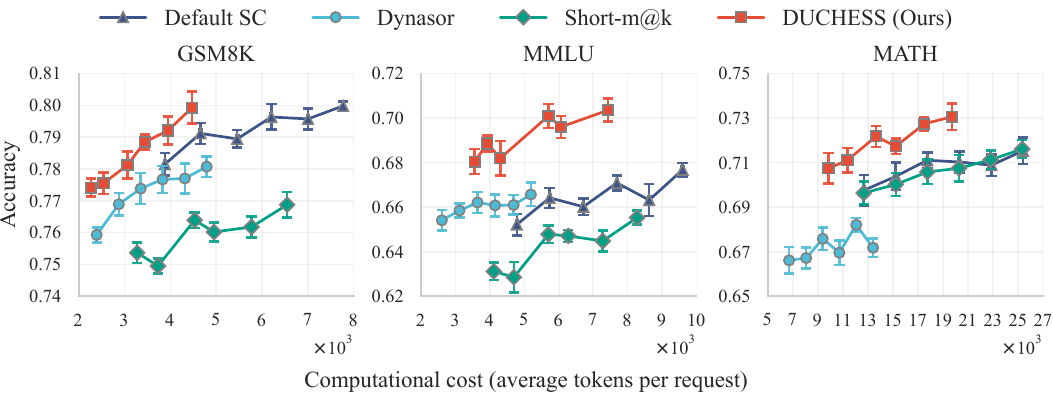}
  \caption{\label{fig:token-accuracy} \textbf{\name reduces computational cost without compromising accuracy.} Cost-accuracy tradeoffs with branches per request varying between 5 and 10 inclusively. Markers within a curve correspond to increasing number of branches per request from left to right. The results are averaged from 10 trials, with error bars representing 95\% confidence intervals.}
\end{figure}

\textbf{\name dominates the cost-accuracy Pareto frontier across benchmarks} (Figure~\ref{fig:token-accuracy}). \name consistently achieves higher accuracy at the same token budget and lower token usage at the same accuracy, strictly outperforming baselines in all of our experiments. For example, while matching default SC's max accuracy (within 0.1\% tolerance), \name reduces token usage by $42\%$, $63\%$, and $46\%$ respectively across three benchmarks, demonstrating significant inference cost savings. Notably, at $c=10$, \name achieves $2.7\%$ and $1.5\%$ higher accuracy than Default SC on MMLU and MATH datasets, respectively. This is due to \name's early termination improving per-branch accuracy rates, which we discuss more in~\secref{sec:analysis}.




Existing works that only apply early termination to reasoning branches reduce accuracy compared to Default SC, highlighting the advantage of branch orchestration. Dynasor consistently reduces tokens but drops max accuracy by $1\%$ to $5\%$ across all tested datasets and number of branches per request. This shows that intermediate answer consistency cannot always indicate correctness. Short-m@k reduces max accuracy across all $c$s by $2\%$ and $3\%$ on GSM8K and MMLU datasets. We observe that shorter branches in several requests are more likely to be incorrect due to underthinking, which the Short-m@k is biased towards. In addition, Short-m@k has negligible token usage reduction (0.2\% at $c=10$) on the MATH dataset because branch lengths are similar in most requests, leaving little room for pruning.


\begin{table}[t]
\centering
\small
\setlength{\tabcolsep}{6pt}
\renewcommand{\arraystretch}{1.12}
\begin{tabular*}{\linewidth}{@{\extracolsep{\fill}} l *{3}{S} *{3}{S} *{3}{S} @{}}
\toprule
& \multicolumn{3}{c}{\textbf{GSM8K}}
& \multicolumn{3}{c}{\textbf{MMLU}}
& \multicolumn{3}{c}{\textbf{MATH}} \\
\cmidrule(lr){2-4}\cmidrule(lr){5-7}\cmidrule(lr){8-10}
\textbf{Method}
& {\textbf{Mean}} & {\textbf{P50}} & {\textbf{P95}}
& {\textbf{Mean}} & {\textbf{P50}} & {\textbf{P95}}
& {\textbf{Mean}} & {\textbf{P50}} & {\textbf{P95}} \\
\midrule
\textbf{Latency (s)} & & & & & & & & & \\
Default SC   & 54.1 & 40.8 & 156.1 & 82.8 & 62.7 & 215.3 & 137.7 & 117.3 & 319.5 \\
Short-m@k    & 15.7 & 10.0 & 44.3 & 23.9  & 14.7  & 72.2 & 69.9 & 60.0  & 182.9 \\
DUCHESS      & 10.5  & 7.7  & 24.7  & 17.0  & 9.2   & 66.7  & 58.8  & 48.8  & 154.4  \\
\midrule
\textbf{TTFT (s)} & & & & & & & & & \\
Default SC   & 34.2 & 16.1 & 134.5 & 59.7 & 39.9 & 191.0 & 88.2 & 63.7 & 260.3 \\
Short-m@k    & 5.1 & 0.070 & 27.4 & 10.3  & 0.072  & 52.3 & 33.8 & 10.8  & 142.1 \\
DUCHESS      & 2.8  & 0.068  & 14.4  & 6.7  & 0.067   & 43.7  & 27.6  & 2.21  & 118.6  \\
\bottomrule
\end{tabular*}
\caption{\label{tab:lat_ttft}  \textbf{\name reduces request latecies and TTFTs.} Mean/P50/P95 latencies and time-to-first-token (TTFT) in seconds, measured from one run using 10 branches per request. The accuracies of Default SC/Short-m@k/\name are 0.80/0.77/0.80 on GSM8K, 0.67/0.66/0.70 on MMLU, 0.71/0.70/0.73 on MATH. Requests follow a Poisson arrival process with mean rates of 2, 2, and 1 requests/min for GSM8K, MMLU, and MATH, respectively.}
\vspace{-5pt}
\end{table}



\noindent\textbf{\name reduces Mean/P50/P95 latencies and TTFTs in pooled request serving.} Table~\ref{tab:lat_ttft} reports latency and TTFT numbers with 10 branches per request. Across three datasets, \name reduces mean, P50, and P95 latencies by 57-81\%, 58-85\%, and 52-84\%, respectively. \name achieves even larger reductions in TTFTs, with 69-92\%, 96-99\%, and 54-89\% in mean, P50, and P95 cases, as a result of significantly lower queueing delays. We include Short-m@k measurements for reference, although its accuracies are often unmatched to Default SC (shown in Figure~\ref{fig:token-accuracy}).

\begin{figure}[t]
  \centering
  \includegraphics[width=\linewidth]{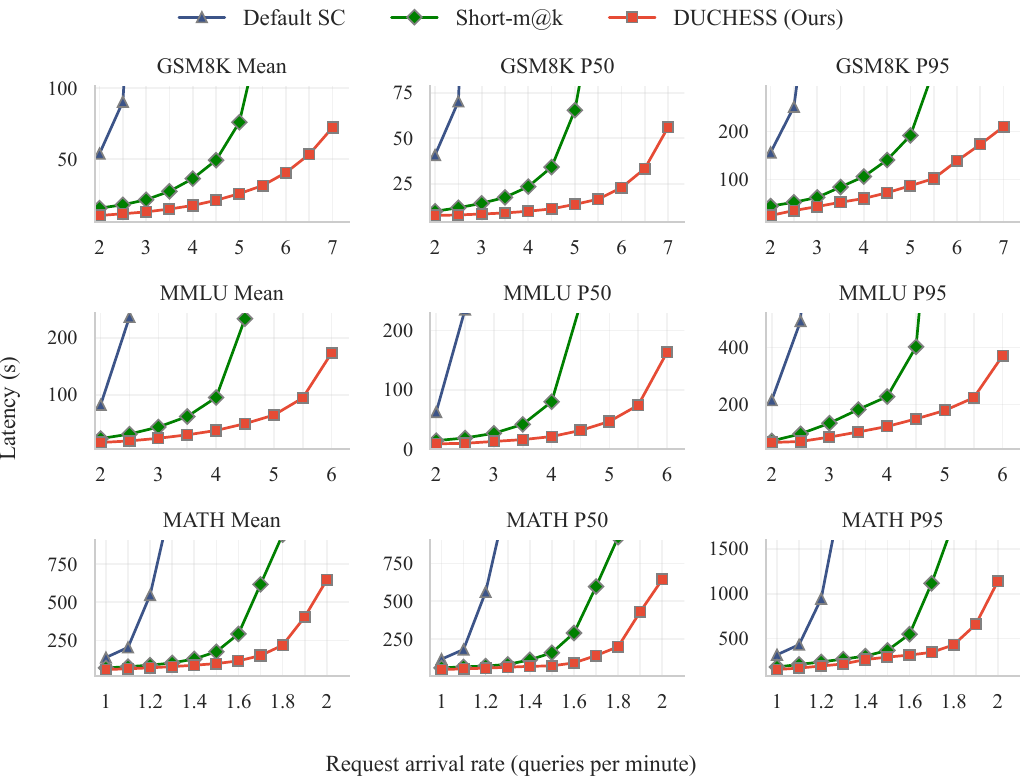}
  \caption{\label{fig:latency_qpm} \textbf{\name is robust to bursts of requests.} Mean, P50, and P95 latencies as a function of request arrival rate (10 branches per request).}
\end{figure}


\noindent\textbf{\name is robust to bursts of requests.} Figure~\ref{fig:latency_qpm} shows the mean, P50, and P95 latencies as a function of request arrival rate, using up to $2\text{-}3.5\times$ higher QPM than rates in Table~\ref{tab:lat_ttft}. Other settings remain unchanged. \name delivers the lowest latencies across reported request loads. With matched mean latency to Default SC at 1.5 QPM, \name can process $3\times$, $2.5\times$, and $1.6\times$ more requests across three datasets, demonstrating robustness to bursts of requests. We also observe that Short-m@k also substantially reduces latency compared to Default SC (although at smaller magnitudes than \name), highlighting the latency inflation caused by straggler branches.

\subsection{Inter-request complexity-aware scheduling.}

\begin{figure}[t]
\centering
  \includegraphics[width=0.8\linewidth]{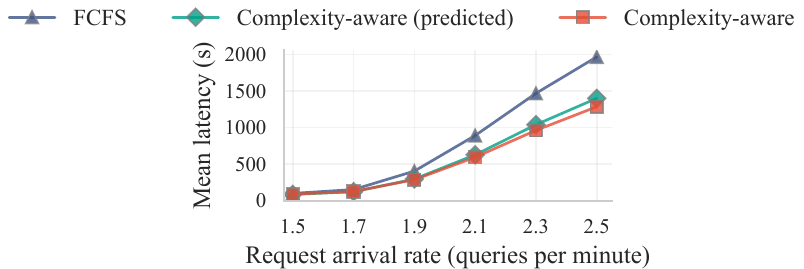}
  \caption{\label{fig:latency-scheduling} \textbf{Complexity-aware request scheduling reduces mean latency compared to FCFS.} Mean latency as a function of request arrival rate using FCFS, complexity-aware using predicted and actual request complexities, on the MATH dataset (10 branches per request).}
\end{figure}

Figure~\ref{fig:latency-scheduling} shows the mean latency on the MATH dataset as a function of request arrival rates, for \name using FCFS and complexity-aware scheduling with predicted and labeled complexities, using 10 branches per request.

\noindent\textbf{Complexity-aware scheduling lowers mean latency relative to FCFS, with larger gains at higher request rates.} Across six QPM values from $1.5$ to $2.5$, compexity-aware scheduling reduces mean latency by $10.4\text{-}34.7\%$ using actual request complexities, and by $14.2\text{-}29.7\%$ using predicted complexities, compared to FCFS. As QPM increases, request queues grow, amplifying the benefits of complexity-aware scheduling.

Our results show that
complexity-aware scheduling outperforms FCFS even with predicted complexities, making it applicable when true complexities are unavailable at test time, provided labeled examples are available to train the predictor.

\subsection{\label{sec:analysis}Analysis}

\noindent\textbf{Accuracy of early-terminated CoT branches.} Terminating a CoT early can reduce accuracy if triggered too soon; conversely, it can sometimes improve accuracy by preventing the LLM from overthinking~\citep{hassid2025dontoverthinkitpreferring}. We compared the accuracies of early terminated branches relative to their uninterrupted accuracies with maximum 4,096 tokens per branch, for 10 branches per request across all three datasets.

Early termination has a negligible impact on GSM8K; only 0.1\% of branches that would be correct if uninterrupted are mistakenly early-terminated. This likely reflects GSM8K's lower difficulty. In contrast, 16.4\% and 7.8\% of branches are mistakenly early-terminated on MMLU and MATH, respectively. However, \name still matches or exceeds Default SC accuracy because: (1) early termination corrects 20.8\% and 22.4\% of branches that would otherwise be incorrect (overthinking or incomplete), and (2) majority voting across branches within each request mitigates individual branch errors.


\begin{table}[t]
\centering
\small
\setlength{\tabcolsep}{6pt}
\renewcommand{\arraystretch}{1.12}
\begin{tabular}{lccc}
      \toprule
      \textbf{Component} & \textbf{GSM8K} & \textbf{MMLU} & \textbf{MATH} \\
      \midrule
      Prediction        & 0.0055 (0.07\%) & 0.0022 (0.02\%) & 0.0053 (0.02\%) \\
      Answer Extraction & 0.28 (3.65\%)   & 0.18 (1.73\%)   & 0.28 (0.89\%) \\
      LLM Inference     & 7.50 (96.3\%)   & 10.12 (98.2\%)  & 30.97 (99.1\%) \\
      \arrayrulecolor{gray}\midrule
      \arrayrulecolor{black}
      \textbf{Total (s/request)} & \textbf{7.8 (100\%)} & \textbf{10.3 (100\%)} & \textbf{31.3 (100\%)} \\
      \bottomrule
    \end{tabular}
  \captionsetup{type=table}
  \vspace{-5pt}
  \caption{\label{tab:component-overhead} \textbf{\name adds negligible orchestration overhead.} The average time in seconds that each component of intra-request orchestration spends per request across three datasets, using 10 branches per requst. The percentages represent relative weights per component within total request processing time.}
\end{table}


\noindent\textbf{Per-component overhead.} \name introduces two additional GPU-consuming steps to standard LLM inference at intra-request level: (1) predicting per-branch correctness at each token interval; and (2) extracting final answers from early-terminated branches. Both lie on the critical path, since \name relies on their output to make decisions at the branch and request levels. They must therefore be lightweight to preserve low end-to-end latency. Table~\ref{tab:component-overhead} breaks down \name's mean per-request processing time by component, using 10 branches per request. The overhead of predicting branch correctness is insignificant across three datasets: between $0.02\%$ and $0.07\%$, due to the lightweight predictor design. We additionally report that the predictor's GPU memory footprint is as small as 8.7 to 40.3 MBs, making it feasible to co-locate on the same device as the LLM (hence omitting potential overhead of transmitting activations). The answer extraction overhead is controlled within $0.89\%$ to $3.65\%$, as it costs only 10 tokens per early-terminated branch.

%% file: sections/conclusion.tex
\section{Conclusion and Limitations}

We have presented \name, a novel approach that significantly reduces computation and latency for multi-branch LLM reasoning without compromising accuracy. Our key contribution is an intra-request branch orchestration policy. We obtain per-branch correctness probabilities using a lightweight MLP model directly based on LLM activations. The policy dynamically decides an action per branch, including early termination, selective branch-out, and continuation, based on the predictions. The policy terminates the request in its entirety once sufficient final answers are collected to prevent stragglers from inflating completion time.  \name also includes an optional complexity-aware scheduler that prioritizes easier requests, given predicted difficulty levels. We evaluate \name using three reasoning datasets: GSM8K, MMLU, and MATH, on the DeepSeek-R1-Distill-Llama-8B model. At matched accuracy to Default Self-Consistency, \name reduces computational cost by generating $42\%\text{-}63\%$ fewer tokens. Using vLLM as the inference engine, \name reduces mean, P50, and P95 latencies by $52\text{-}85\%$ and TTFTs by $54\text{-}99\%$.

\noindent\textbf{Limitations.} Our intra-request branch orchestration terminates a request based on collected answers. However, these answers have dependencies (e.g., some may come from branches that share the same ancestor before branch-out). Moreover, answers are collected when branches are stopped at varying correctness predictions. We aim to explore answer aggregation and request-termination strategies that account for this information. Our evaluation of complexity-aware inter-request scheduling is limited to the MATH dataset. As a next step, we aim to extend complexity-aware scheduling to datasets without difficulty labels.

%% file: sections/appendix.tex
\newpage
\section{Appendix}

\subsection{\label{sec:eval-system-setting}System settings for evaluation.}

\noindent\textbf{Branch correctness predictor.} The MLP for per-branch correctness prediction has an input layer of dimension 4,096, corresponding to the hidden size of DeepSeek-R1-Distill-LLama-8B, followed by a LayerNorm, then two hidden fully-connected (FC) layers, and finally an output layer of dimension 1. The hidden layer sizes are (2048, 1024) for GSM8K and MATH datasets, and (512, 256) for MMLU. Each hidden layer is followed by a ReLU activation and a dropout layer with 0.15 probability.

The MLPs are optimized using AdamW with an initial learning rate of $3\times10^{-4}$ and 0.1 weight decay. We adjust the learning rate over epochs using StepLR scheduler with step size 5 and gamma 0.8. We use Binary Cross Entropy loss function, appropriate for classification tasks. We train the model for up to 15 epochs, and keep the model weights from the epoch with highest validation accuracy.

We use activations from layer 14 of Distill-R1-Llama as input to the MLP for all three datasets. Our exhaustive search over all layers shows that Layer 14 yields highest validation accuracies for GSM8K (Figure~\ref{fig:pred_acc_by_layer}). Due to time and resource constraints, we do not profile all layers for other two datasets.

\noindent\textbf{Intra-request branch orchestration hyperparameter settings.} We tune \name hyperparameters based on the cost-accuracy tradeoffs on validation set. We then use the same settings for latency evaluations. In general, \name performs similarly on validation and test sets.

$\tau$ (early termination threshold): we use the $70^{th}$ percentile of all validation sample predictions for GSM8K, and $80^{th}$ percentile for MATH and MMLU. GSM8K's lower $\tau$ reflects its lower difficulty compared to other datasets, as \name is able to tolerate more mistakenly early terminated branches while not reducing overall accuracy.

$\lambda$ (sampling temperature for selective branch-out): we use $1$ for GSM8K, and $0.8$ for MATH and MMLU. We did not test values outside of $0.8-1$ ranges. The selected $\lambda$ values keep branch selection weights close or identical to predicted correctness, with slightly increased weight on high-confidence branches for MATH and MMLU, reflecting their advanced difficulties.

$\alpha, \beta$ (request-level termination): we use $(0.6, 0.8)$ for $\alpha, \beta$ values on GSM8K and MATH datasets, and $(0.4, 1)$ on MMLU.

$i$ (interval): we set $i=80$ for MMLU and MATH, and a smaller $i=16$ for GSM8K due to shorter LLM responses.

\noindent\textbf{Request complexity predictor.} Our request complexity predictor uses activations from layer 14 of R1-Distill-Llama. We did not test other layers, although they may further improve prediction accuracy and mean latencies. The MLP predictor has an input layer with 4,096 dimension, matching R1-Distill-Llama's hidden sizes, followed by a LayerNorm, then three hidden FC layers with 2048, 1024, 512 dimensions. Each FC layer is followed by a BatchNorm1d, then GeLU activation, and a dropout layer. The dropout rates are 0.15 for first two, and 0.075 for the last FC layer. The output from the last FC layer is passed to another dropout with 0.045 rate, and finally an output layer with dimension 5, corresponding to the number of unique difficulty labels in MATH dataset. The model is optimized using AdamW with a consine annealing learning rate scheduler, and 0.1 label smoothing for regularization. Overall, we applied extensive regularization efforts, but the validation and test accuracies remain around $0.4$. Nevertheless, the predicted complexities are sufficient to reduce mean latencies compared to FCFS.

\noindent\textbf{vLLM inference settings.} We use vLLM v0.7.3 with \texttt{enable-prefix-caching} set to true; this ensures KV reuse for forked branches and answer extraction from early-terminated branches. We use matched sampling parameters to~\citep{fu2024efficient}, with 0.6, 0.95, and 1 for temperature, top p, and n, respectively.

\subsection{\label{sec:complexity-prediction-acc}Confusion matrix for request complexity predictor.}

\begin{figure}[h]
  \centering
  \includegraphics[width=0.32\linewidth]{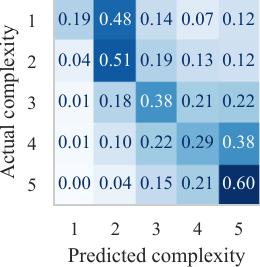}
  \caption{Confusion matrix for predicted request complexities on MATH dataset.}
  \label{fig:math-difficulty-cm}
\end{figure}

Figure~\ref{fig:math-difficulty-cm} shows the confusion matrix for predicted request complexities. Although the predictor has relatively low overall and balanced accuracies (0.42 and 0.39), we observe that it is able to differentiate easier and harder requests well: for example, $81\%$ level-1 requests have predicted complexity levels $\leq 3$, while $81\%$ level-5 requests have predictions $> 3$. Our evaluation demonstrates that these predictions guide \name to prioritize easier requests and reduce mean latencies.

